\documentclass[paper]{ieice}
\usepackage[fleqn]{amsmath}
\usepackage{newtxtext}
\usepackage[varg]{newtxmath}
\usepackage{bm}
\usepackage{mathtools}
\usepackage{booktabs}
\usepackage{makecell}
\usepackage{cite}

\field{D}
\title{Localized Kernel Projection Outlyingness: A Two-Stage Approach for Multi-Modal Outlier Detection}
\authorlist{%
  \authorentry{Akira Tamamori}{m}{AIT}\MembershipNumber{1512145}
  \affiliate[AIT]{The author is with the Department the
    Computer Science, Aichi Institute of Technology, Aichi, 470-0392, Japan.}
}
\received{2015}{1}{1}
\revised{2015}{1}{1}

\begin{document}
\maketitle

\begin{summary}
  This paper presents Two-Stage LKPLO, a novel multi-stage outlier
  detection framework that overcomes the coexisting limitations of
  conventional projection-based methods: their reliance on a fixed
  statistical metric and their assumption of a single data
  structure. Our framework uniquely synthesizes three key concepts:
  (1) a generalized loss-based outlyingness measure (PLO) that
  replaces the fixed metric with flexible, adaptive loss functions
  like our proposed SVM-like loss; (2) a global kernel PCA stage to
  linearize non-linear data structures; and (3) a subsequent local
  clustering stage to handle multi-modal distributions. Comprehensive
  5-fold cross-validation experiments on 10 benchmark datasets, with
  automated hyperparameter optimization, demonstrate that Two-Stage
  LKPLO achieves state-of-the-art performance. It significantly
  outperforms strong baselines on datasets with challenging structures
  where existing methods fail, most notably on multi-cluster data
  (Optdigits) and complex, high-dimensional data
  (Arrhythmia). Furthermore, an ablation study empirically confirms
  that the synergistic combination of both the kernelization and
  localization stages is indispensable for its superior
  performance. This work contributes a powerful new tool for a
  significant class of outlier detection problems and underscores the
  importance of hybrid, multi-stage architectures.
\end{summary}

\section{Introduction}
Outlier detection remains a pivotal challenge in data science, with
profound implications across diverse fields from financial fraud
prevention to the discovery of astronomical anomalies. Among the
various approaches, the projection pursuit paradigm, epitomized by the
Stahel-Donoho Outlyingness (SDO)~\cite{stahel1981, donoho1992}, offers
a robust and intuitive framework. It posits that a multivariate
outlier will also manifest as an outlier in at least one suitably
chosen one-dimensional projection. This principle, valued for its
robustness to high dimensionality and non-elliptical data
distributions, has inspired a family of methods, including the
computationally efficient Random Projection Depth
(RPD)~\cite{zuo2003}, which form a cornerstone of modern robust
statistics.

Despite the elegance of this paradigm, its classical formulation
suffers from two key limitations. First, its reliance on a fixed
statistical metric, typically the median absolute deviation (MAD),
restricts its ability to adapt to diverse data structures. Second, the
assumption of a single, global data structure makes it vulnerable to
multi-modal distributions. In the context of this paper, we use the
term ``multi-modal'' in its statistical sense to describe data
distributions characterized by multiple distinct peaks or clusters
(e.g., disjoint groups of inliers), rather than data originating from
different sensory modalities (e.g., text and images). When classical
projection-based methods encounter such multi-cluster structures,
distinct clusters of normal data can be easily misinterpreted as
anomalies.  Recent research has attempted to address these issues
independently. On one hand, methods like Kernel Outlier Detection
(KOD)~\cite{tamamori2023apsipa, dagidir2025} have introduced
sophisticated, data-driven ensembles of projection directions to
handle complex structures, yet they still rely on the fixed MAD-based
metric. On the other hand, our previous work on Weighted Random
Projection Outlyingness (WRPO)~\cite{tamamori2023} demonstrated the
effectiveness of a clustering-based localization strategy for handling
multi-modality, but remained confined to the original data space and
fixed metric. A unified framework capable of addressing both
limitations simultaneously has yet to be established.

In this paper, we propose a unified framework, namely Two-Stage
Localized Kernel Projection-based Loss Outlyingness (Two-Stage LKPLO),
that addresses both of these fundamental challenges
simultaneously. Our approach begins by redefining outlyingness not as
a fixed statistical distance but as a flexible loss maximization
problem. This conceptual shift is inspired by the recent work of
Castellanos and Mozharovskyi~\cite{castellanos2025}, who established a
connection between data depth and statistical risk, is here extended
and adapted to the projection pursuit paradigm. We refer to this
generalized formulation as Projection-based Loss Outlyingness
(PLO). This enables the replacement of the rigid MAD-based metric with
more powerful, task-adaptive loss functions such as the proposed
SVM-like loss. To further handle multi-modality, we introduce a
two-stage architecture. In the first stage, a global kernel PCA is
performed to construct a feature space in which non-linear structures
are linearized. In the second stage, clustering is applied within this
feature space to identify local data structures, and the PLO score is
then computed locally within each cluster’s context, weighted by
cluster size.

This synergistic design results in a method that is both powerful and
flexible. Our main contributions are threefold: (1) we introduce a
unified framework that integrates a generalized, loss-based measure of
outlyingness with a localized, kernel-based projection strategy; (2)
we develop a concrete and computationally efficient algorithm,
Two-Stage LKPLO, incorporating a novel SVM-like loss function; and (3)
through extensive experiments on a wide range of benchmark datasets,
we demonstrate that our method achieves state-of-the-art performance,
particularly on datasets with complex multi-cluster structures such as
Optdigits and Vowels, where previous methods have struggled. We
further show that our approach not only outperforms existing methods
in its areas of strength but also remains highly competitive across
diverse data types, validating the effectiveness of our integrated
design.

The remainder of this paper is organized as follows. Section 2 reviews
related work in the fields of projection-based, kernel-based, and
clustering-based outlier detection. Section 3 details the formulation
of our proposed Two-Stage LKPLO framework. Section 4 presents a
comprehensive experimental evaluation, including qualitative analysis
on synthetic datasets and quantitative comparisons on real-world
benchmarks. Finally, Section 5 concludes the paper and discusses
future directions.

\section{Related Work}

Our work lies at the intersection of three major research streams in
outlier detection: projection-based methods, kernel-based methods, and
clustering-based methods.

\subsection{Projection-based Outlyingness}
\label{sec:rw_projection}
The foundation of our approach lies in the projection pursuit
paradigm, pioneered by Stahel and Donoho’s Outlyingness
(SDO)~\cite{stahel1981, donoho1992}. This family of methods searches
for one-dimensional projections where outliers are most apparent. To
improve computational efficiency, Random Projection Depth
(RPD)~\cite{zuo2003} introduced the use of a random subset of
projection directions. Although robust and scalable, these classical
methods remain fundamentally limited by their reliance on a fixed
statistical metric (MAD) and their assumption of a single global data
structure, both of which hinder performance on multi-modal data.

\subsection{Kernel-based Methods}
\label{sec:rw_kernel}
To handle non-linear data structures, kernel-based approaches have
been widely explored. The One-Class SVM (OCSVM)~\cite{Scholkopf2001}
learns a separating hyperplane in a high-dimensional feature
space. More relevant to our work, recent studies have explicitly
combined the kernel trick with projection pursuit. For example, Kernel
RPD (KRPD)~\cite{tamamori2023apsipa} computes RPD within a
kernel-induced feature space. Building on this idea, Kernel Outlier
Detection (KOD)~\cite{dagidir2025} introduced a sophisticated ensemble
of data-driven projection directions to enhance detection
performance. However, while these methods effectively address
non-linearity, they still inherit the limitation of a fixed MAD-based
metric from their projection-based predecessors.

\subsection{Clustering-based and Local Methods}
\label{sec:rw_clustering}
A third stream of research focuses on the local context of data
points. Methods such as Local Outlier Factor (LOF)~\cite{Breunig2000}
measure the deviation of a point’s local density from that of its
neighbors. More structurally, clustering-based approaches such as
Clustering-Based LOF (CBLOF)~\cite{Zengyou2003} and our previous work
on Weighted Random Projection Outlyingness (WRPO)~\cite{tamamori2023}
first partition the data into clusters and then evaluate outlyingness
based on both cluster size and the distance to each cluster
centroid. These methods are effective for multi-modal data but often
operate in the original data space, making them vulnerable to complex
non-linear structures (the ``curse of dimensionality’’), and they
typically lack the flexibility of projection-based evaluation.

\subsection{Our Contribution in Context}
\label{sec:rw_contribution}
The literature reveals a clear trend toward hybrid approaches;
however, a method that simultaneously addresses the limitations of a
fixed metric, non-linearity, and multi-modality has remained
elusive. Our proposed Two-Stage LKPLO, detailed in the following
section, is designed to fill this gap by uniquely synthesizing the
principles of all three research streams into a single, coherent
framework. To the best of our knowledge, no existing work has achieved
this level of integration.

\section{Proposed Method: Two-Stage Localized Kernel PLO}
The proposed method, Two-Stage Localized Kernel PLO (Two-Stage LKPLO),
addresses complex outlier structures by integrating three key
concepts: a generalized measure of outlyingness, a non-linear mapping
via the kernel method, and a localization strategy based on
clustering.

The cornerstone of our framework is the Projection-based Loss
Outlyingness (PLO), a generalization of the classical projection
pursuit paradigm. While inspired by the conceptual link between data
depth and statistical risk established by Castellanos and
Mozharovskyi~\cite{castellanos2025}, our PLO framework diverges in two
fundamental respects. First, whereas their ``risk'' is formulated as a
minimization problem to find the most central point (i.e., to measure
``depth''), our notion of outlyingness is formulated as a loss
maximization problem, designed to identify the projection where a
point appears most anomalous. Second, their framework operates in the
full multivariate space, whereas ours is specifically adapted to the
projection pursuit paradigm, evaluating loss in one-dimensional
projected space. This adaptation enables us to leverage the extensive
literature on projection directions and makes our approach
computationally tractable for high-dimensional outlier detection.

The Two-Stage LKPLO algorithm, detailed below, implements this PLO
concept within a powerful multi-stage architecture comprising a global
feature-learning stage and a local scoring stage.

\subsection{Stage 1: Global Non-linear Feature Learning via Kernel PCA}
Given a training dataset
$X = \{\bm{x}_{1}, \bm{x}_{2}, \ldots, \bm{x}_{N}\} \subset
\mathbb{R}^{d}$, the first step is to map the data into a
high-dimensional feature space $\mathcal{F}$ using a kernel function
$k(\cdot, \cdot)$. We employ the Radial Basis Function (RBF) kernel,
$k(\bm{x}_{i}, \bm{x}_{j}) = \exp(-\gamma \lVert\bm{x}_{i} -
\bm{x}_{j}\rVert^{2})$.

The core of this stage is to perform Kernel Principal Component
Analysis (KPCA)~\cite{Scholkopf1998kpca} on the entire dataset
$X$. This process begins with the computation of the $N \times N$ Gram
matrix $\bm{K}$ and its centered version $\bar{\bm{K}}$. By solving
the eigenvalue problem for $\bar{\bm{K}}$, we obtain a mapping to a
$q$-dimensional kernel feature space. Each data point
$\bm{x}_{i} \in X$ is thus transformed into a $q$-dimensional feature
vector $\bm{f}_{i} \in \mathbb{R}^{q}$, yielding a new representation
$\bm{F} = \{\bm{f}_{1}, \bm{f}_{2}, \ldots, \bm{f}_{N}\}$, where the
non-linear structures of the inliers are expected to be linearized. A
new test point $\bm{x}_{\text{new}}$ is transformed into its feature
vector $\bm{f}_{\text{new}}$ using the same learned mapping.

\subsection{Stage 2: Localization and Scoring}
This stage comprises two main steps: clustering in the feature space
and local outlyingness scoring.  This second stage is the core of our
approach to handle multi-modal data. By first partitioning the global
feature space into $K$ clusters, we can treat each cluster as a
unimodal local structure. The subsequent outlyingness scoring is then
performed within the local context of each cluster, thus avoiding the
problem of misinterpreting distinct normal clusters as anomalies.

\subsubsection{Clustering in the Feature Space}
We apply a clustering algorithm, such as \textit{k}-means, to the
feature representation $\bm{F}$ to partition it into $K$ clusters. Let
$\mathcal{C} = \{c_1, c_2, \ldots, c_K\}$ denote the set of
clusters. For each cluster $c_{k} \in \mathcal{C}$, we compute and
store its constituent feature vectors $\bm{F}_k \subset \bm{F}$, its
size $N_{k}= |\bm{F}_{k}|$, and its centroid
$\boldsymbol{\mu}_{k} = \frac{1}{N_k} \sum_{\bm{f} \in \bm{F}_k}
\bm{f}$.

\subsubsection{Local Projection-based Loss Outlyingness}
The final outlyingness score for a test point $\bm{x}_{\text{new}}$ is
computed based on its local context.  First, its feature vector
$\bm{f}_{\text{new}}$ is assigned to the nearest cluster $c_k$ by
minimizing the Euclidean distance to the cluster centroids
$\{\boldsymbol{\mu}_{1}, \dots, \boldsymbol{\mu}_{K}\}$.

The local outlyingness is then defined as the maximum loss over a set
of projection directions $U$, computed in the coordinate system
centered at the cluster centroid.  Let
$\bm{f}'_{\text{new}} = \bm{f}_{\text{new}} -
\boldsymbol{\mu}_k$ and
$\bm{F}'_k = \{ \bm{f} - \boldsymbol{\mu}_k \mid \bm{f}
\in \bm{F}_k \}$ be the centered feature vectors. The local
outlyingness is given by:
\begin{equation}
  \text{local-score}(\bm{f}_{\text{new}}) = \max_{\bm{u} \in U} \mathcal{L}(\bm{u}, \bm{f}'_{\text{new}}, \bm{F}'_k)
    \label{eq:local_score}
\end{equation}
where $\mathcal{L}(\cdot)$ is a chosen loss function. In this work, we
investigate two loss functions:
\begin{itemize}
\item \textbf{Robust Z-score Loss:} This is the classical
  Stahel-Donoho outlyingness measure, defined as:
  \begin{equation}
    \mathcal{L}_{\text{RZ}}(\bm{u}, \bm{f}'_{\text{new}}, \bm{F}'_k) = \frac{|\bm{u}^{\top} \bm{f}'_{\text{new}} - \text{median}(\bm{u}^{\top} \bm{F}'_k)|}{\text{MAD}(\bm{u}^{\top} \bm{F}'_k)},
        \label{eq:loss_rz}
      \end{equation}
      where $\text{median}(\cdot)$ is the median operator, and
      $\text{MAD} (\cdot)$ is the Median Absolute Deviation, defined
      for a univariate sample $\{\bm{z}_{i}\}$ as
      \begin{equation}
        \text{MAD}(\{\bm{z}_{i}\}) = 1.4826\cdot \text{median}_{i}(|\bm{z}_{i} - \text{median}_{j}(\bm{z}_{j})|).
         \nonumber
    \end{equation}
  \item \textbf{SVM-like Loss:} Inspired by the concept of margins in
    Support Vector Machines, we propose a loss that measures the
    deviation from a robustly estimated boundary of the inlier
    cluster. Instead of scaling by the MAD, it uses a scaled version
    of the MAD as a soft margin or radius, $\rho_k$. The loss is
    defined as:
    \begin{equation}
        \mathcal{L}_{\text{SVM}}(\bm{u}, \bm{f}'_{\text{new}}, \bm{F}'_k) = \max(0, |\bm{u}^{\top} \bm{f}'_{\text{new}}| - \rho_k),
        \label{eq:loss_svm}
    \end{equation}
    where the radius
    $\rho_k = c \cdot \text{MAD}(\bm{u}^{\top} \bm{F}'_k)$, with $c$
    being a hyperparameter. This formulation leverages the same robust
    scale estimator as the Z-score loss but employs it in a
    fundamentally different way: it defines a boundary rather than
    normalizing a distance. We chose this SVM-like formulation for two
    primary reasons. First, inspired by Support Vector Machines, it
    aims to find a boundary rather than a central point, which we
    hypothesized is more effective for the non-Gaussian distributions
    found in the kernel space. Second, its formulation using
    $\text{max}(0, \ldots)$ results in a loss of zero for points deep
    inside the inlier region, leading to a form of sparsity where only
    boundary points influence the outlyingness score, which can
    improve robustness.
\end{itemize}
We also considered other loss functions, such as the logistic
loss. For brevity, we focus on the two primary variants in the main
text and provide a detailed formulation and evaluation of the logistic
loss variant in Appendix A.

The set of projection directions $U$ is constructed as an ensemble of
data-driven directions, including Random, Basis Vector, One-Point, and
Two-Points types, inspired by the work of Da\u{g}{\i}d{\i}r et
al.~\cite{dagidir2025}.

Finally, following the principle of our previous work on
WRPO~\cite{tamamori2023}, the final score is weighted by the inverse
of the cluster size to penalize points belonging to smaller,
potentially anomalous clusters:
\begin{equation}
    \text{Score}(\bm{x}_{\text{new}}) = \frac{1}{N_k} \cdot \text{local-score}(\bm{f}_{\text{new}})
    \label{eq:final_score}
  \end{equation}

\subsection{Computational Complexity}
We briefly discuss the computational complexity of the proposed
Two-Stage LKPLO. The main bottleneck lies in the training phase,
specifically the computation of the kernel matrix ($O(N^2 d)$) and its
subsequent eigen-decomposition ($O(N^3)$), where $N$ is the number of
training samples and $d$ is the original dimensionality.  The inference
phase is relatively efficient, as it mainly involves kernel
evaluations against the training data $(O(MNd))$ and a fast projection
search. While the cubic complexity in $N$ can be a challenge for very
large datasets, this can be effectively mitigated by employing kernel
approximation techniques such as the Nystr{\"o}m method, as
demonstrated in~\cite{Williams2000}. This suggests that our framework is
adaptable to large-scale settings, though a detailed analysis is left
for future work.

\section{Experiments}

In this section, we conduct a comprehensive set of experiments to
empirically evaluate the performance of our proposed method, Two-Stage
LKPLO. We aim to answer the following key questions:

\begin{itemize}
\item How does Two-Stage LKPLO perform compared to standard and
  state-of-the-art outlier detection methods across a wide range of
  datasets?
\item Under what data structures does our proposed method show its
  strengths and limitations?
\item How do the different components of our method, such as the
  choice of loss function and the localization strategy, contribute to
  its performance?
\end{itemize}

\subsection{Experimental Setup}
\label{sec:exp_setup}

\paragraph{Datasets}
We evaluate our method on two fronts: an intuitive qualitative
analysis using 2D synthetic datasets, and a large-scale quantitative
evaluation using publicly available benchmark datasets.

For the qualitative analysis, we use three challenging synthetic
datasets designed to highlight specific failure modes of existing
algorithms:
\begin{itemize}
\item \textbf{Three Gaussians}: A multi-modal dataset where inliers
  form three distinct clusters.
\item \textbf{Inside--Outside}: A challenging dataset specifically
  designed for this study, featuring inliers in a ring structure
  with outliers both inside and outside the ring.
\item \textbf{Moons}: A standard benchmark for non-linear separability.
\end{itemize}

For the quantitative evaluation, we selected 10 benchmark datasets
from the ODDS library~\cite{Rayana2016}.  The characteristics of these
benchmark datasets are summarized in Table~\ref{tab:ODDS}, which
details the number of samples, dimensionality, and the percentage of
anomalies for each dataset. As shown in the table, we selected
datasets covering a diverse range of complexities. Importantly, all
selected benchmark datasets consist of continuous numerical features,
ensuring the direct applicability of the Euclidean distance-based RBF
kernel in our framework. Furthermore, datasets such as Optdigits,
Vowels, and Letter are specifically included as they are well-known
canonical examples of data exhibiting distinct multi-cluster (or
multi-modal) structures, which are central to evaluating our
localization strategy.

\begin{table}[t]
  \caption{
    Information of benchmark dataset~\cite{Rayana2016}: dataset name,
    number of samples, number of feature dimensions, and outlier percentages.}
  \label{tab:ODDS}
  \centering
  \begin{tabular}{l c c c}
    \toprule
    Dataset & \#samples & \#dims. & outliers (\%)\\
    \midrule
    Arrhythmia & 452 & 274 & 14.6\% \\
    Cardio & 1831 & 21 & 9.61\% \\
    Glass & 214 & 9 & 4.2\% \\
    Ionosphere & 351 & 33 & 35.9\% \\
    Letter & 1600 & 32 & 6.25\% \\
    Optdigits & 5216 & 64 & 2.88\%\\
    Pima & 768 & 8 & 35\% \\
    Vertebral & 240 & 6 & 12.5\% \\
    Vowels & 1456 & 12 & 3.43\% \\
    WBC & 378 & 30 & 5.56\%\\
    \bottomrule
  \end{tabular}
\end{table}

\paragraph{Comparison Methods}
We compare the proposed Two-Stage LKPLO with a suite of strong
baseline and state-of-the-art methods. The evaluated methods include:
\begin{itemize}
\item \textbf{Standard non-kernel methods}: \textit{k}-nearest
  neighbors (KNN)~\cite{Ramaswamy2000}, LOF~\cite{Breunig2000} and
  Kernel Density Estimation (KDE)~\cite{Latecki2007}.
\item \textbf{Standard kernel-based methods}:
  OCSVM~\cite{Scholkopf2001} and KPCA~\cite{Hoffman2007}.
\item \textbf{Standard clustering-based method}: CBLOF~\cite{Zengyou2003}.
\item \textbf{Standard tree-based method}: Isolation Forest (IForest)~\cite{IForest2008}.
\item \textbf{Related projection-based methods}: This category
  includes methods that are conceptually direct predecessors to our
  proposed framework, allowing for a direct evaluation of our
  architectural contributions.
  \begin{itemize}
  \item \textbf{RPD}~\cite{zuo2003}: The classical projection-based
    method, serving as the foundational baseline.
  \item \textbf{WRPO}~\cite{tamamori2023}: An extension of RPD that
    introduces a \textit{localization} strategy in the original data
    space. Although a clustering-based method in its own right (as
    discussed in Sec.\ref{sec:rw_clustering}), we include it here to
    specifically evaluate the impact of the localization component.
  \item \textbf{KRPD}~\cite{tamamori2023apsipa}: An extension of
    RPD that introduces a \textit{kernel mapping} but remains a
    global method.
  \item \textbf{KOD}~\cite{dagidir2025}: A recent state-of-the-art
    method that enhances kernelized projection pursuit with a
    sophisticated ensemble of data-driven directions. While modern and
    hybrid, it retains a global scoring approach, providing a
    benchmark to evaluate the necessity of our two-stage localization.
  \end{itemize}
\end{itemize}
For our proposed method, we evaluate its two loss function variants:
Two-Stage LKPLO (SVM-like) and (Robust-Z).

\paragraph{Evaluation Protocol}
To ensure a robust and fair comparison, we employ a 5-fold stratified
cross-validation protocol for each dataset. We use
\verb|StratifiedKFold| from scikit-learn, which preserves the
percentage of samples for each class (inlier/outlier) in each fold. A
fixed random seed (\verb|random_state|=42) is applied to all
data-splitting processes to ensure the reproducibility of our
experiments. The final reported score for each method is the mean and
standard deviation of the ROC AUC across the 5 folds.

\paragraph{Hyperparameter Optimization}
For all methods, including our proposed approach, we perform automatic
hyperparameter optimization using the Optuna
framework~\cite{optuna_2019}. In each fold, the designated training
data is further divided into a training subset (75\%) and a validation
subset (25\%), again using stratification and a fixed random seed. For
each method, Optuna is executed for 50 trials to identify the
hyperparameter configuration that maximizes the ROC AUC on the
validation subset. This rigorous procedure ensures that all methods
are evaluated under fair and consistent tuning conditions, allowing
for a comparison of their optimal potential.

While the search spaces for baseline methods followed standard
practices, the key hyperparameter ranges for our proposed Two-Stage
LKPLO were set as follows. For the SVM-like variant:
\begin{itemize}
\item Number of clusters, $K$: integers in $[2, 30]$
\item RBF kernel parameter, $\gamma$: log-uniform in $[10^{-4}, 10^{1}]$
\item Number of principal components, $q$: integers in $[5, 30]$
\item SVM-like loss hyperparameter, $c$: uniform in $[1.0, 5.0]$
\end{itemize}
The ranges for the Robust-Z variant were set similarly.

\paragraph{Implementation Details}
All experiments were conducted on a machine equipped with an Intel
Core i9-9900K CPU, 64 GB of RAM, and running Ubuntu 24.04 LTS. The
proposed method and all comparison experiments were implemented in
Python 3.12. We utilized several key libraries, including scikit-learn
1.7.1, PyOD 2.0.5, Optuna 4.4.0, NumPy 2.2.6, and Matplotlib 3.10.3.

\subsection{Evaluation on Synthetic Datasets}

\begin{figure*}[t]
  \centering
  
  \begin{minipage}[b]{0.32\textwidth}
    \centering
    \includegraphics[width=\textwidth]{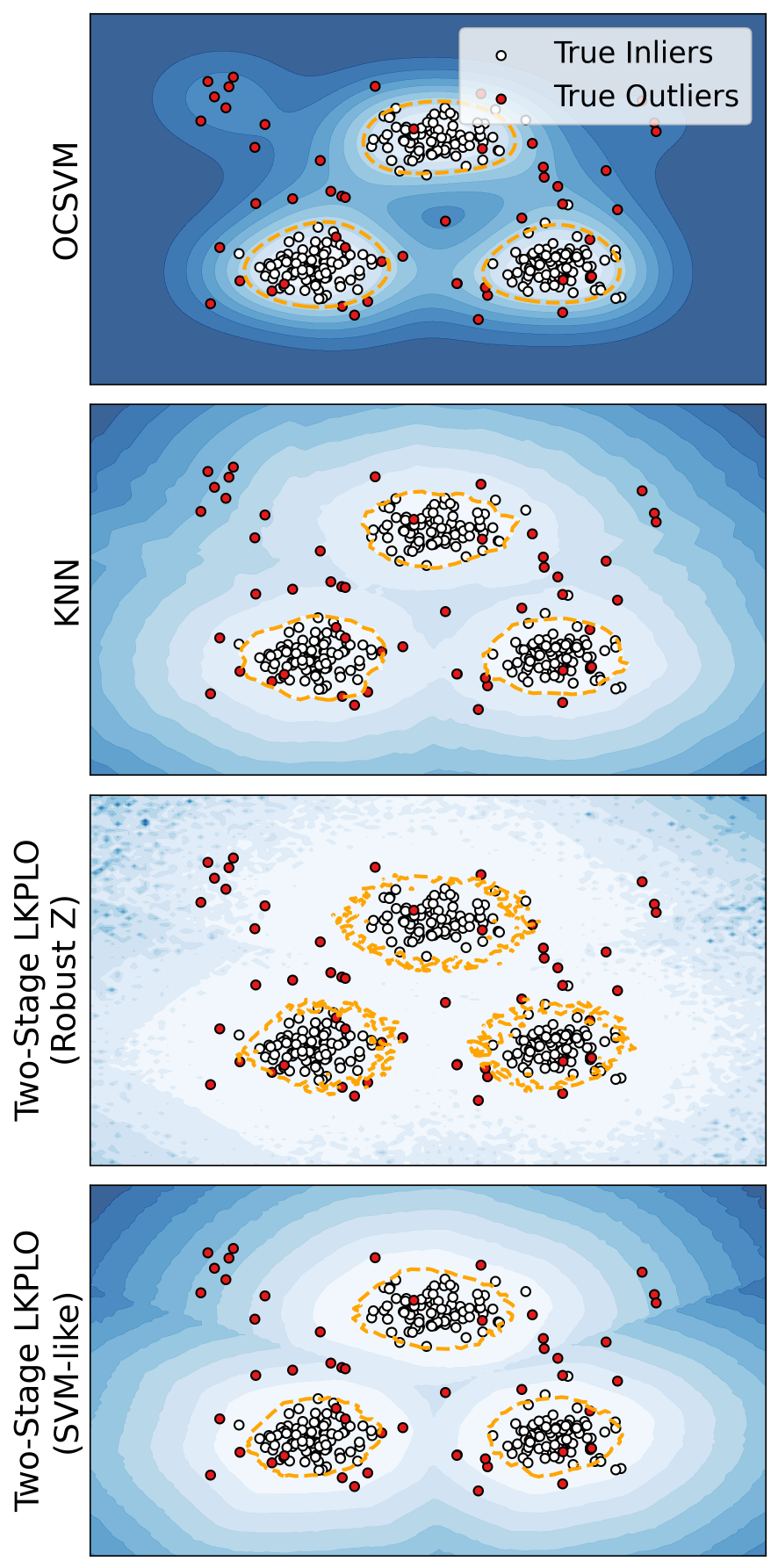}
    \centerline{(a) Three Gaussians}
  \end{minipage}
  \hfill
  \begin{minipage}[b]{0.32\textwidth}
    \centering
    \includegraphics[width=\textwidth]{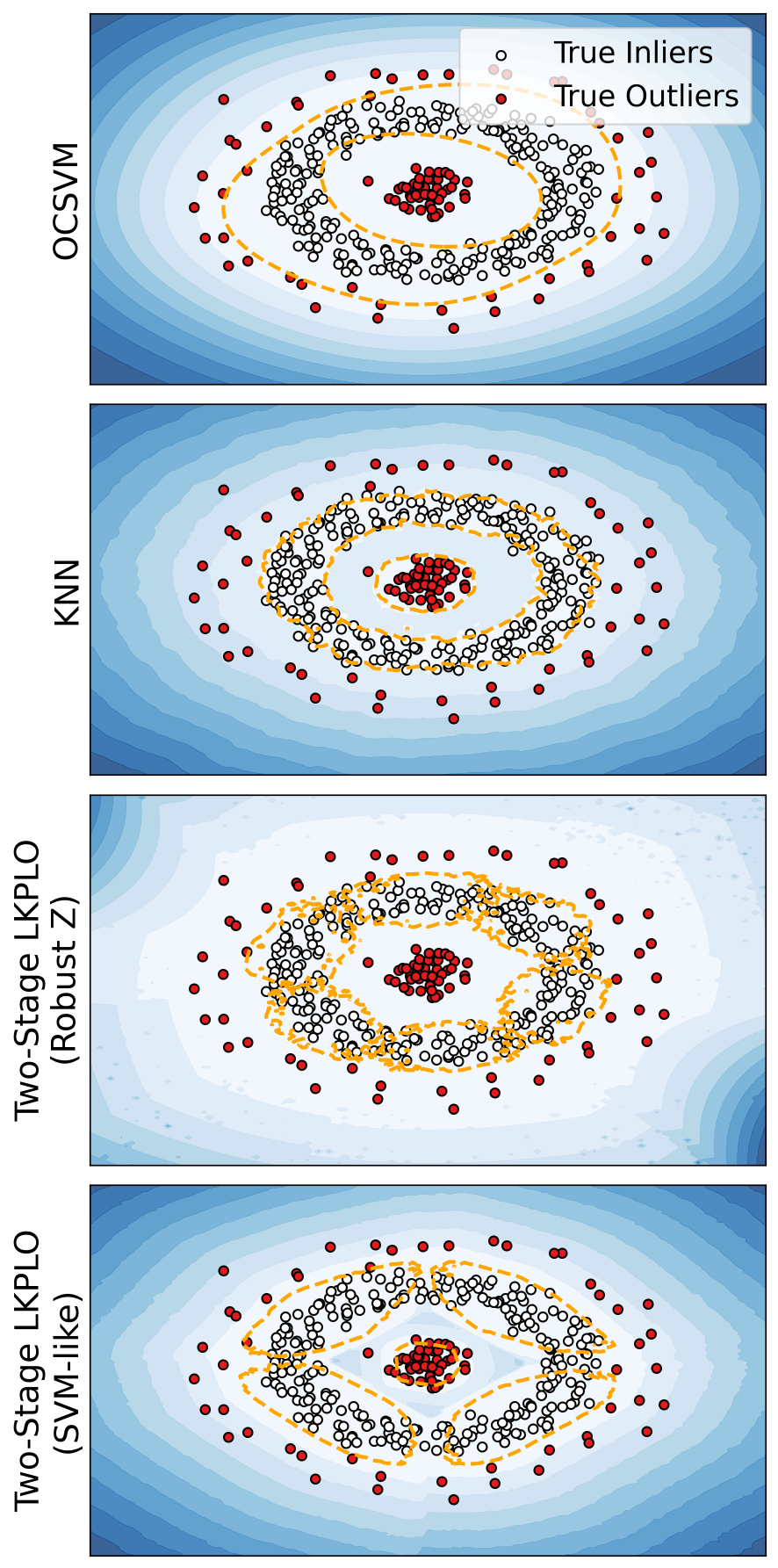}
    \centerline{(b) Inside--Outside}
  \end{minipage}
  \hfill
  \begin{minipage}[b]{0.32\textwidth}
    \centering
    \includegraphics[width=\textwidth]{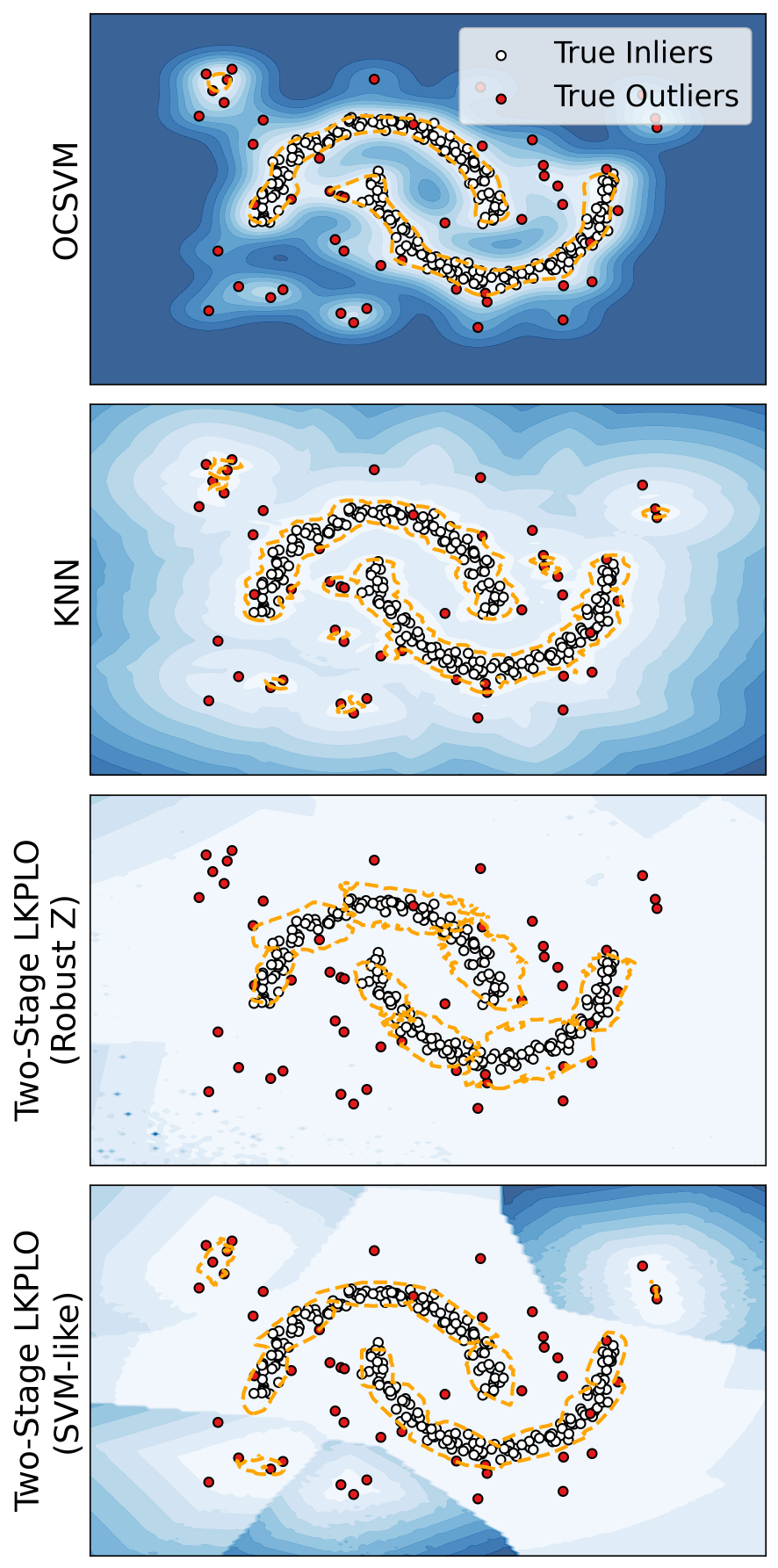}
    \centerline{(c) Moons}
  \end{minipage}
  
  \caption{Comparison of optimized decision boundaries on three
    synthetic datasets. In each subfigure (a)-(c), rows from top to
    bottom correspond to different outlier detection algorithms:
    OCSVM, KNN, Two-Stage LKPLO (Robust Z), and Two-Stage LKPLO
    (SVM-like). The background color map represents the outlier score,
    where darker shades indicate higher outlyingness. The orange
    dashed line shows the decision boundary separating inliers from
    outliers. The orange dashed line shows the decision boundary
    separating inliers from outliers.}
  \label{fig:plot_toydata}
\end{figure*}

Figure~\ref{fig:plot_toydata} presents a visual comparison of the
decision boundaries learned by OCSVM, KNN, and the two variants of our
proposed Two-Stage LKPLO, after hyperparameter optimization using
Optuna on the three synthetic datasets introduced in Section 4.1. This
qualitative analysis reveals the distinctive characteristics and
capabilities of each algorithm.  In these visualizations, the orange
dashed lines enclose the regions classified as normal (inliers) by
each model, while the background color maps represent the continuous
outlier scores (darker blue indicates higher outlyingness).

The top two panels of each subfigure show the results of the standard
baselines. On the \textbf{Three Gaussians} dataset, a relatively
straightforward multi-modal scenario, both OCSVM and KNN successfully
place their decision boundaries around the three distinct inlier
clusters. However, their underlying limitations become apparent upon
closer inspection of the score distributions. OCSVM, which attempts to
learn a single global boundary, models the three clusters as a single,
irregular distribution; the background color map reveals that the
regions between the clusters are not assigned sufficiently high
anomaly scores. On the \textbf{Inside--Outside} dataset, while OCSVM's
global nature allows it to correctly exclude the dense central cluster
from the normal region, it struggles to accurately capture the hollow
structure of the ring, resulting in an overly broad decision boundary
that incorrectly classifies parts of the inner void as
normal. Conversely, KNN, a local method, forms an overly complex and
fragmented boundary on \textbf{Moons} due to its sensitivity to local
density variations. On \textbf{Inside--Outside}, this reliance on
local density leads to a critical failure: KNN incorrectly classifies
the dense central cluster of outliers as normal, enclosing it within
an inner boundary. These results demonstrate the distinct failure
modes and limitations inherent in purely global and purely local
approaches when faced with complex data structures.

The third panel of each subfigure displays the results of our
Two-Stage LKPLO framework utilizing the classical Robust Z-score
loss. Aided by the two-stage architecture, this variant successfully
isolates the distinct inlier clusters in the \textbf{Three Gaussians}
dataset. However, its limitations become apparent when faced with more
complex geometries. On the \textbf{Inside--Outside} dataset, while it
manages to encompass the inliers, the resulting outer boundary is
distorted and wider than necessary, incorporating regions of empty
space. On the \textbf{Moons} dataset, the piecewise nature of the
local boundaries is evident, yet the overall shape struggles to
smoothly adapt to the non-convex curvature. Instead, it forms a series
of somewhat rigid, convex-like segments that, while capturing the
majority of inliers, lack the precision to tightly follow the data
manifold. These observations empirically demonstrate that while
localization is a powerful step, relying on a traditional,
distance-based metric (MAD) within the kernel-induced space is often
insufficient to learn highly precise, complex boundaries.

The bottom panel highlights the full potential as well as a specific
vulnerability of our proposed Two-Stage LKPLO (SVM-like) framework:
\begin{itemize}
\item \textbf{Three Gaussians}: It learns tight, well-separated
  boundaries for each inlier cluster, matching the success of the
  baselines and confirming the effectiveness of the clustering stage.
\item \textbf{Moons}: It learns a tight, non-convex boundary that
  intricately follows the underlying data manifold. The
  piecewise-linear nature of this boundary reflects the mechanism of
  the local projection search, effectively approximating a complex
  shape through a collection of simpler, locally adaptive
  decisions. However, a close inspection reveals occasional, small
  isolated regions (or ``islands'') classified as normal in the
  periphery, which can lead to minor false negatives (misclassifying
  outliers as inliers). This artifact highlights a potential
  vulnerability of our two-stage approach: if the clustering stage
  inadvertently identifies isolated noise points or empty regions as
  distinct micro-clusters, the subsequent local scoring mechanism may
  erroneously assign them low anomaly scores within their restricted
  local context.
\item \textbf{Inside--Outside}: While our method captures the overall
  circular structure of the inlier ring through a combination of
  stable local boundaries (unlike the more fragmented boundary of
  KNN), it reveals a critical limitation regarding dense, clustered
  anomalies (micro-clusters). Similar to KNN, our Two-Stage LKPLO
  incorrectly encloses the central outlier cluster within a normal
  boundary. This failure highlights a vulnerability in the
  localization stage (Stage 2): if a dense group of outliers is
  identified as a distinct cluster by the \textit{k}-means algorithm,
  our method evaluates them within their own local context, resulting
  in erroneously low anomaly scores. This demonstrates that while our
  hybrid architecture excels at modeling complex inlier manifolds
  (like \textbf{Moons}), it remains susceptible to clustered anomalies
  that masquerade as valid structural components of the data.
\end{itemize}

In summary, this visual analysis underscores that the superior
performance of our proposed method stems from the synergistic
interplay among its key components: the kernel mapping linearizes the
data; the two-stage localization effectively handles multi-modality;
and the flexible SVM-like loss function enables the learning of
adaptive, non-convex boundaries.

\subsection{Performance Comparison on Benchmark Datasets}
\begin{table*}[!ht]
  \centering
  \caption{ROC AUC scores (mean $\pm$ std.\ dev.) of outlier detection
    methods across 10 benchmark datasets, based on 5-fold
    cross-validation. The best performance for each dataset is
    highlighted in bold. (Part 1: Comparison with Baseline Methods)}
  \label{tab:results_part1}
    \begin{tabular}{lccccccc} 
    \toprule
    \textbf{Datasets} & \textbf{KNN} & \textbf{LOF} & \textbf{KDE} & \textbf{IForest} & \textbf{CBLOF} & \textbf{OCSVM} & \textbf{KPCA} \\
    \midrule
    Arrhythmia & 0.775 $\pm$ 0.080 & 0.766 $\pm$ 0.084 & 0.742 $\pm$ 0.063 & 0.810 $\pm$ 0.030 & 0.781 $\pm$ 0.080 & 0.769 $\pm$ 0.072 & 0.771 $\pm$ 0.074 \\
    Cardio     & 0.892 $\pm$ 0.019 & 0.736 $\pm$ 0.029 & 0.951 $\pm$ 0.013 & 0.926 $\pm$ 0.021 & 0.854 $\pm$ 0.087 & 0.948 $\pm$ 0.014 & 0.941 $\pm$ 0.013 \\
    Glass      & 0.812 $\pm$ 0.092 & 0.710 $\pm$ 0.270 & 0.727 $\pm$ 0.183 & 0.693 $\pm$ 0.130 & 0.727 $\pm$ 0.164 & 0.732 $\pm$ 0.173 & 0.749 $\pm$ 0.160 \\
    Ionosphere & 0.930 $\pm$ 0.022 & 0.903 $\pm$ 0.019 & 0.937 $\pm$ 0.021 & 0.850 $\pm$ 0.035 & 0.869 $\pm$ 0.074 & 0.934 $\pm$ 0.024 & \textbf{0.940 $\pm$ 0.026} \\
    Letter     & 0.910 $\pm$ 0.035 & 0.891 $\pm$ 0.032 & 0.896 $\pm$ 0.046 & 0.663 $\pm$ 0.019 & 0.805 $\pm$ 0.032 & \textbf{0.926 $\pm$ 0.019} & 0.921 $\pm$ 0.025 \\
    Optdigits  & 0.405 $\pm$ 0.022 & 0.586 $\pm$ 0.067 & 0.563 $\pm$ 0.037 & 0.710 $\pm$ 0.055 & 0.731 $\pm$ 0.054 & 0.554 $\pm$ 0.073 & 0.585 $\pm$ 0.053 \\
    Pima       & \textbf{0.724 $\pm$ 0.028} & 0.663 $\pm$ 0.033 & 0.716 $\pm$ 0.026 & 0.684 $\pm$ 0.032 & 0.686 $\pm$ 0.037 & 0.698 $\pm$ 0.041 & 0.723 $\pm$ 0.035 \\
    Vertebral  & 0.344 $\pm$ 0.071 & 0.512 $\pm$ 0.104 & 0.378 $\pm$ 0.094 & 0.395 $\pm$ 0.063 & 0.412 $\pm$ 0.055 & 0.560 $\pm$ 0.104 & 0.489 $\pm$ 0.114 \\
    Vowels     & \textbf{0.962 $\pm$ 0.028} & 0.934 $\pm$ 0.020 & 0.876 $\pm$ 0.097 & 0.790 $\pm$ 0.052 & 0.928 $\pm$ 0.039 & 0.894 $\pm$ 0.108 & 0.922 $\pm$ 0.068 \\
    WBC        & 0.937 $\pm$ 0.048 & 0.924 $\pm$ 0.053 & 0.939 $\pm$ 0.045 & 0.927 $\pm$ 0.052 & 0.951 $\pm$ 0.040 & 0.940 $\pm$ 0.046 & 0.932 $\pm$ 0.045 \\
    \midrule
    \textbf{Average} & 0.769 $\pm$ 0.203 & 0.762 $\pm$ 0.144 & 0.772 $\pm$ 0.187 & 0.745 $\pm$ 0.180 & 0.774 $\pm$ 0.169 & 0.795 $\pm$ 0.139 & 0.810 $\pm$ 0.174 \\
    \bottomrule
    \end{tabular}
  \vspace{5mm}
  \centering
  \caption{(Continued) ROC AUC scores. (Part 2: Comparison with Projection-based Methods and Our Proposed Method)}
  \label{tab:results_part2}
  \begin{tabular}{lcccccc} 
    \toprule
    \textbf{Datasets} & \textbf{RPD} & \textbf{WRPO} & \textbf{KRPD} & \textbf{KOD} &\textbf{\thead{Two-Stage LKPLO \\ (ours, Robust-Z)}} & \textbf{\thead{Two-Stage LKPLO \\ (ours, SVM-like)}} \\
    \midrule
     Arrhythmia & 0.762 $\pm$ 0.076 & 0.758 $\pm$ 0.086 & 0.765 $\pm$ 0.084 & 0.782 $\pm$ 0.069 & 0.794 $\pm$ 0.052 & \textbf{0.813 $\pm$ 0.058} \\
    Cardio     & 0.939 $\pm$ 0.019 & 0.931 $\pm$ 0.036 & \textbf{0.972 $\pm$ 0.008} & 0.958 $\pm$ 0.017 & 0.846 $\pm$ 0.070 & 0.937 $\pm$ 0.011 \\
    Glass      & 0.746 $\pm$ 0.105 & 0.810 $\pm$ 0.114 & 0.720 $\pm$ 0.087 & 0.481 $\pm$ 0.220 & 0.878 $\pm$ 0.070 & \textbf{0.881 $\pm$ 0.105} \\
    Ionosphere & 0.863 $\pm$ 0.011 & 0.823 $\pm$ 0.082 & 0.934 $\pm$ 0.020 & 0.877 $\pm$ 0.014 & 0.862 $\pm$ 0.078 & 0.862 $\pm$ 0.048 \\
    Letter     & 0.653 $\pm$ 0.047 & 0.841 $\pm$ 0.052 & 0.873 $\pm$ 0.023 & 0.853 $\pm$ 0.013 & 0.818 $\pm$ 0.067 & 0.832 $\pm$ 0.033 \\
    Optdigits  & 0.572 $\pm$ 0.074 & 0.871 $\pm$ 0.007 & \textbf{0.926 $\pm$ 0.035} & 0.794 $\pm$ 0.188 & 0.888 $\pm$ 0.085 & 0.922 $\pm$ 0.027 \\
    Pima       & 0.695 $\pm$ 0.039 & 0.704 $\pm$ 0.028 & 0.691 $\pm$ 0.018 & 0.697 $\pm$ 0.084 & 0.041 $\pm$ 0.029 & 0.711 $\pm$ 0.024 \\
    Vertebral  & 0.369 $\pm$ 0.083 & 0.386 $\pm$ 0.113 & \textbf{0.651 $\pm$ 0.112} & 0.634 $\pm$ 0.102 & 0.650 $\pm$ 0.049 & 0.629 $\pm$ 0.136 \\
    Vowels     & 0.790 $\pm$ 0.041 & 0.924 $\pm$ 0.027 & 0.784 $\pm$ 0.107 & 0.882 $\pm$ 0.031 & 0.904 $\pm$ 0.016 & 0.906 $\pm$ 0.029 \\
    WBC        & 0.919 $\pm$ 0.054 & 0.931 $\pm$ 0.047 & 0.928 $\pm$ 0.077 & 0.927 $\pm$ 0.059 & 0.937 $\pm$ 0.052 & \textbf{0.957 $\pm$ 0.027} \\
    \midrule
    \textbf{Average} & 0.731 $\pm$ 0.183 & 0.798 $\pm$ 0.181 & 0.824 $\pm$ 0.108 & 0.789 $\pm$ 0.080 & 0.823 $\pm$ 0.057 & \textbf{0.843 $\pm$ 0.109} \\
    \bottomrule
  \end{tabular}
\end{table*}

The overall performance comparison across 10 benchmark datasets is
summarized in Tables~\ref{tab:results_part1}
and~\ref{tab:results_part2}.  Each table reports the mean and standard
deviation of ROC AUC scores obtained through five-fold
cross-validation, with the best performance for each dataset
highlighted in bold. Our analysis yields several key insights into the
behavior and effectiveness of the proposed framework.

First, the \textit{Average} scores at the bottom of the tables provide
an overview of the overall trends.  Our proposed Two-Stage LKPLO
(SVM-like) achieves the highest average ROC AUC (0.843) among all nine
evaluated methods, demonstrating strong and consistent performance
across diverse data structures.  This is closely followed by its
conceptual predecessor, KRPD (0.824), underscoring the general
strength of kernelized, projection-based approaches. It is also highly
noteworthy that WRPO (Avg: 0.798), which incorporates localization
into the basic RPD framework (Avg: 0.731), exhibits a substantial
improvement in average performance. This finding strongly suggests
that the localization strategy is a powerful component for handling
multi-modal data, even in the original data space.

\begin{figure*}[t]
  \centering
  \includegraphics[width=0.9\linewidth]{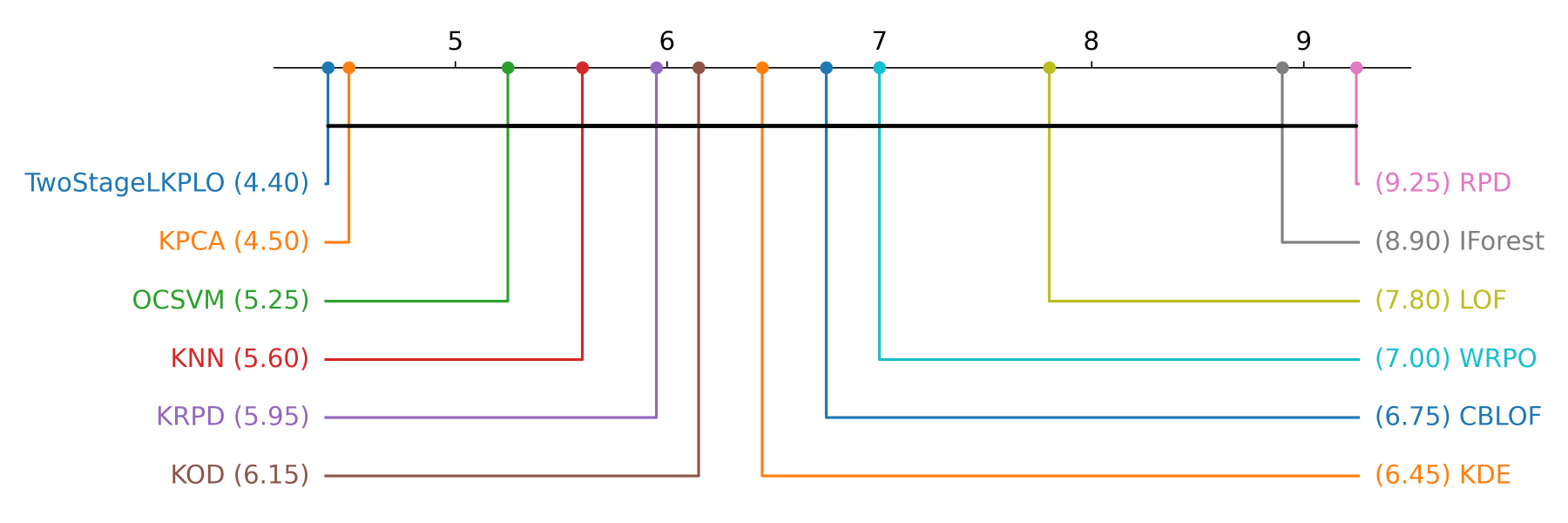}
  \caption{Critical Difference (CD) diagram of the Nemenyi post-hoc
    test ($\alpha=0.05$). The methods are ranked based on their
    average ROC AUC scores across 10 benchmark datasets (lower rank is
    better). Methods connected by a horizontal line are not
    statistically significantly different from one another, whereas
    methods not sharing a line are significantly different. Our
    proposed Two-Stage LKPLO achieves the highest rank.}
  \label{fig:cd_diagram}
\end{figure*}

To assess the statistical significance of the performance differences,
we conducted a Friedman test~\cite{Friedman1940, Demsar2006} on the
ROC AUC scores across all 10 datasets. The test yielded a $p$-value of
0.0397, indicating that the differences between the methods are
statistically significant at $\alpha=0.05$. Subsequently, we performed
a Nemenyi post-hoc test~\cite{Nemenyi1963} and visualized the results
in a Critical Difference (CD) diagram.  As shown in
Fig.~\ref{fig:cd_diagram}, our proposed Two-Stage LKPLO achieves the
best average rank (4.40) among all 12 evaluated methods. It is
noteworthy that the average rank progressively improves from the
foundational RPD (9.25) to WRPO (7.00), KRPD (5.95), and finally to
Two-Stage LKPLO. This visualizes the cumulative impact of our design
choices: localization, kernelization, and loss
generalization. Although the Nemenyi test, which is known for its
conservative nature especially when the number of datasets is small
relative to the number of methods, shows a critical difference line
connecting all models, the clear rank superiority of our method
demonstrates its robustness and competitive edge over standard
baselines such as KNN, LOF, and OCSVM.

The strength of our Two-Stage LKPLO becomes particularly evident on
several challenging datasets:
\begin{itemize}
\item \textbf{Arrhythmia}: On this high-dimensional and complex
  medical dataset, our method achieved the top score (AUC: 0.813),
  significantly outperforming most other methods, including the strong
  baseline IForest. This performance, particularly in a
  high-dimensional and unstructured setting, underscores the power of
  our kernelized and localized approach.
\item \textbf{Glass}: Our framework achieved competitive scores,
  particularly with the SVM-like loss (AUC: 0.881), substantially
  exceeding the next best baseline, KNN (0.812). While the Robust-Z
  variant also performed competitively (AUC: 0.878), the results
  suggest that the two-stage architecture effectively models complex
  inter-feature relationships, regardless of the specific loss
  function. The strong performance of both variants on this dataset,
  which exhibits dense, separated clusters, further validates our
  approach.
\item \textbf{WBC} and \textbf{Optdigits}: Our method demonstrated
  top-tier or highly competitive performance on these canonical
  multi-cluster problems.  The results on Optdigits are particularly
  revealing, as they clearly highlight the distinct contributions of
  localization and kernelization.  Notably, WRPO, which introduces a
  localization strategy to the basic RPD framework, achieves a strong
  AUC of 0.871 (a substantial improvement over the baseline RPD of
  0.572), confirming that localization is a key factor for success on
  multi-cluster data.  However, even higher performance is achieved by
  methods incorporating kernelization. Both our Two-Stage LKPLO
  variants (Robust-Z: 0.888, SVM-like: 0.922) and KRPD (0.926)
  outperform WRPO. This suggests that while localization is crucial,
  first mapping the data into a more separable feature space via
  Kernel PCA enables more effective clustering and local scoring.  By
  integrating both principles, our Two-Stage LKPLO achieves
  state-of-the-art performance, validating our central hypothesis that
  the synergistic combination of global kernel mapping and local
  clustering provides a particularly powerful strategy for modeling
  complex, inherently clustered data structures.
\end{itemize}

It is particularly insightful to compare our Two-Stage LKPLO with KOD,
as both are modern, hybrid, kernel-based projection methods. While KOD
demonstrates strong performance on datasets with a single global
structure like Cardio and Ionosphere, it struggles significantly on
multi-cluster data such as Glass (AUC: 0.481) and Optdigits (AUC:
0.794). In stark contrast, our Two-Stage LKPLO excels on precisely
these datasets (Glass: 0.881, Optdigits: 0.922). This direct
comparison provides compelling evidence that our key contribution, the
two-stage localization strategy, is a critical component for
successfully handling multi-modal data, which is a challenge that even
sophisticated global methods like KOD cannot overcome.

While our proposed method is powerful, the results also confirm the
``No Free Lunch'' principle, as no single method demonstrates optimal
performance across all datasets.
\begin{itemize}
\item \textbf{Cardio} and \textbf{Ionosphere}: On these datasets, KRPD
  and KPCA exhibit superior performance. These datasets likely possess
  global non-linear structures that are sufficiently captured by a
  single kernel mapping, rendering the additional localization step
  unnecessary or potentially even counterproductive due to added
  complexity.
\item \textbf{Vowels} and \textbf{Letter}: On these datasets, which
  are characterized by distinct and well-separated clusters, simpler
  methods such as KNN, OCSVM, and notably WRPO, perform exceptionally
  well.  The strong performance of WRPO (e.g., AUC: 0.912 on Vowels)
  underscores the effectiveness of localization strategies for
  multi-modal data.  The fact that Two-Stage LKPLO achieves a
  comparable score suggests that, for data in which clusters are
  well-separated in the original feature space, the primary
  performance gain arises from localization, while the additional
  complexity introduced by kernel mapping may offer limited further
  benefit.
\end{itemize}

A crucial aspect of our framework lies in the flexibility of its loss
function.  This is clearly demonstrated by the direct comparison
between our two proposed variants in Table~\ref{tab:results_part2}.
While the classical Robust-Z metric performs well on certain datasets
(e.g., achieving a strong AUC of 0.888 on Optdigits), the SVM-like
loss variant consistently attains higher or comparable performance
across most datasets, yielding a superior overall average score (0.843
vs. 0.823). This superiority of the SVM-like loss was further
confirmed against a logistic loss-based variant, which achieved a
lower average AUC of 0.810 (see Appendix A for full results). This
suggests that the latter, which measures deviation from a learned
boundary rather than from a fixed central point, is generally better
suited to the non-Gaussian and complex distributions that arise in the
kernel-induced feature space.  This finding strongly supports the
central thesis of our PLO framework: advancing outlier detection
requires moving beyond fixed statistical metrics toward adaptive,
learnable loss functions.

In summary, the quantitative results demonstrate that our Two-Stage
LKPLO, particularly when equipped with the SVM-like loss, is not a
universal solution but a powerful, state-of-the-art approach for a
significant and challenging class of outlier detection problems.

\subsection{Analysis of Data Structures and Model Behavior}

To understand the varying performance of the algorithms across
different datasets, we visualize the data structures using
PaCMAP~\cite{pacmap}, a dimensionality reduction technique that
effectively preserves both local and global relationships.
Figure~\ref{fig:pacmap_optdigits}, \ref{fig:pacmap_vowels}, and
\ref{fig:pacmap_arrhythmia} show the 2D embeddings for the Optdigits,
Vowels, and Arrhythmia datasets, respectively. These visualizations
provide compelling insights that align well with our quantitative
results presented in Table~\ref{tab:results_part1} and
\ref{tab:results_part2}.

\subsubsection{Case Study 1: Success on Well-Separated Multi-Cluster Data (Optdigits)}

\begin{figure}[t]
  \centering
  \includegraphics[width=\linewidth]{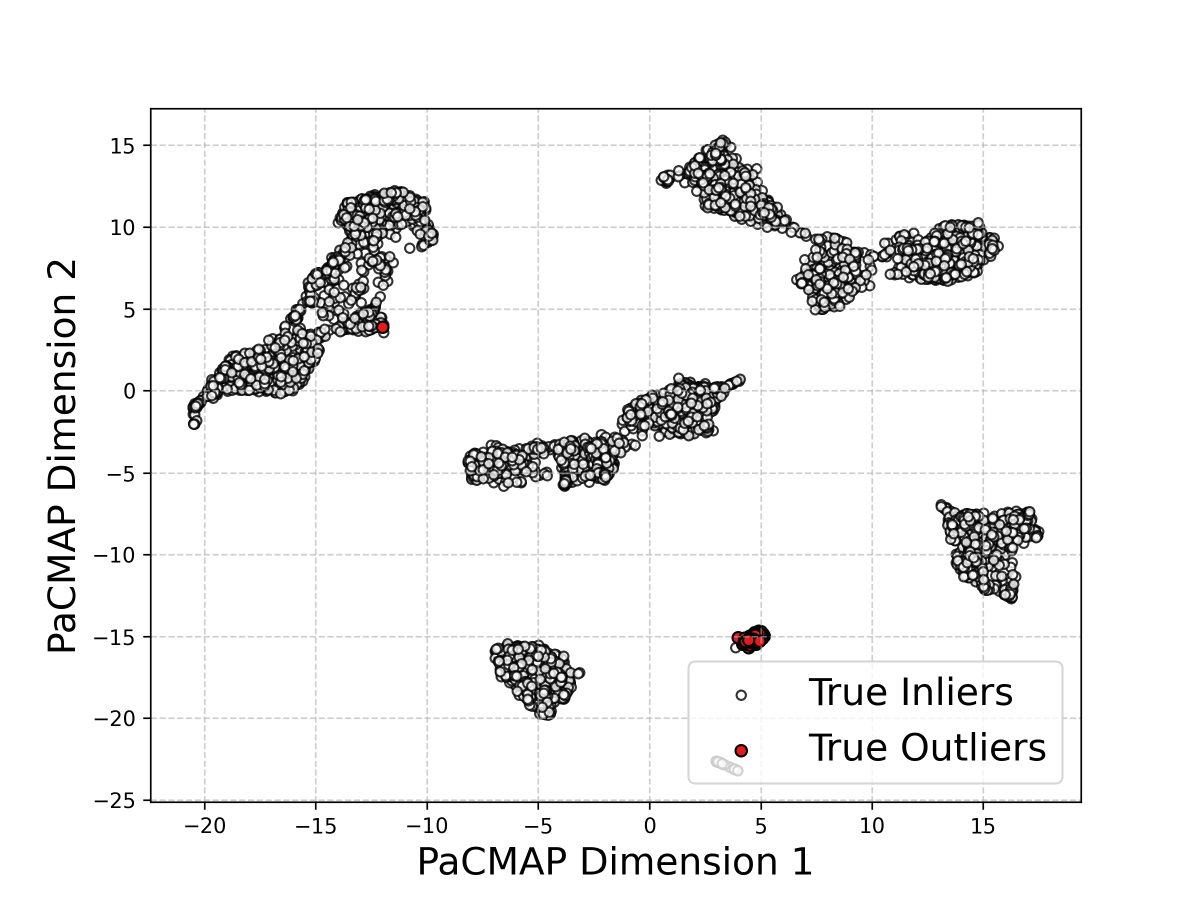}
  \caption{PaCMAP visualization of the Optdigits dataset; It exhibits
    a clear multi-cluster structure. }
  \label{fig:pacmap_optdigits}
\end{figure}

The Optdigits dataset, where our Two-Stage LKPLO achieved its most
decisive performance (AUC: 0.922), exhibits a clearly multi-clustered
structure, as shown in Fig.~\ref{fig:pacmap_optdigits}. The inliers
(white dots) form multiple, dense, and well-separated clusters,
presumably corresponding to different handwritten digits, while the
outliers (red dots) exist as small, isolated groups that do not
conform to any main cluster.

This setting is ideal for our two-stage approach. The global kernel
PCA stage effectively maps the high-dimensional pixel data into a
feature space where these digit clusters become approximately linearly
separable. Subsequently, the clustering stage accurately identifies
these distinct groups, allowing our method to correctly learn the
manifold of each digit class. Outliers, which by definition do not
belong to any of these learned manifolds, are then easily identified
by the local scoring mechanism. In contrast, simpler methods like KNN
struggle in this scenario because outliers can form their own dense
local neighborhoods, leading to false negatives.

\subsubsection{Case Study 2: The Challenge of Locally Dense, Globally Separated Clusters (Vowels)}

\begin{figure}[t]
  \centering
  \includegraphics[width=\linewidth]{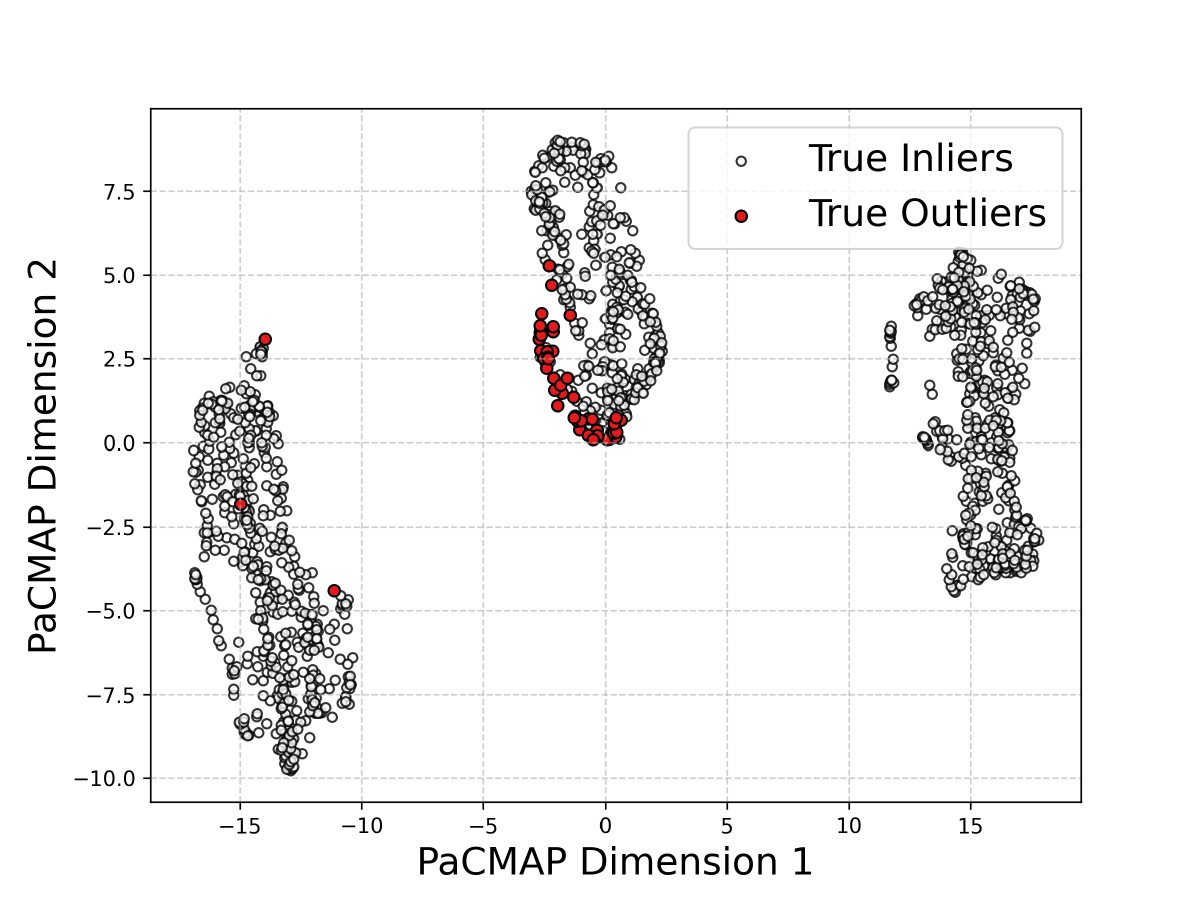}
  \caption{PaCMAP visualization of the Vowels dataset; It is
    characterized by locally dense, globally separated clusters.}
  \label{fig:pacmap_vowels}
\end{figure}

Figure~\ref{fig:pacmap_vowels} visualizes the Vowels dataset, where
KNN demonstrated the strongest performance. The PaCMAP embedding
reveals that the inliers form three distinct, elongated, and highly
dense clusters, while the outliers are sparsely located, often near
the boundaries of these dense regions.

In this context, KNN’s superiority can be attributed to its purely
local nature. Given the high density and clear separation between the
inlier clusters, an outlier’s nearest neighbors are highly likely to
be other outliers or boundary inliers, resulting in a high anomaly
score. Our Two-Stage LKPLO, while achieving a competitive AUC of
0.906, was slightly outperformed. This suggests that although our
localization strategy is effective, the additional complexity of the
kernel mapping and projection search may not be optimal for data where
simple, local density is the most powerful discriminating
feature. Nevertheless, its strong performance validates its ability to
adapt to multi-modal data.

\subsubsection{Case Study 3: Unstructured Data and the Power of a Hybrid Approach (Arrhythmia)}

\begin{figure}[t]
  \centering
  \includegraphics[width=\linewidth]{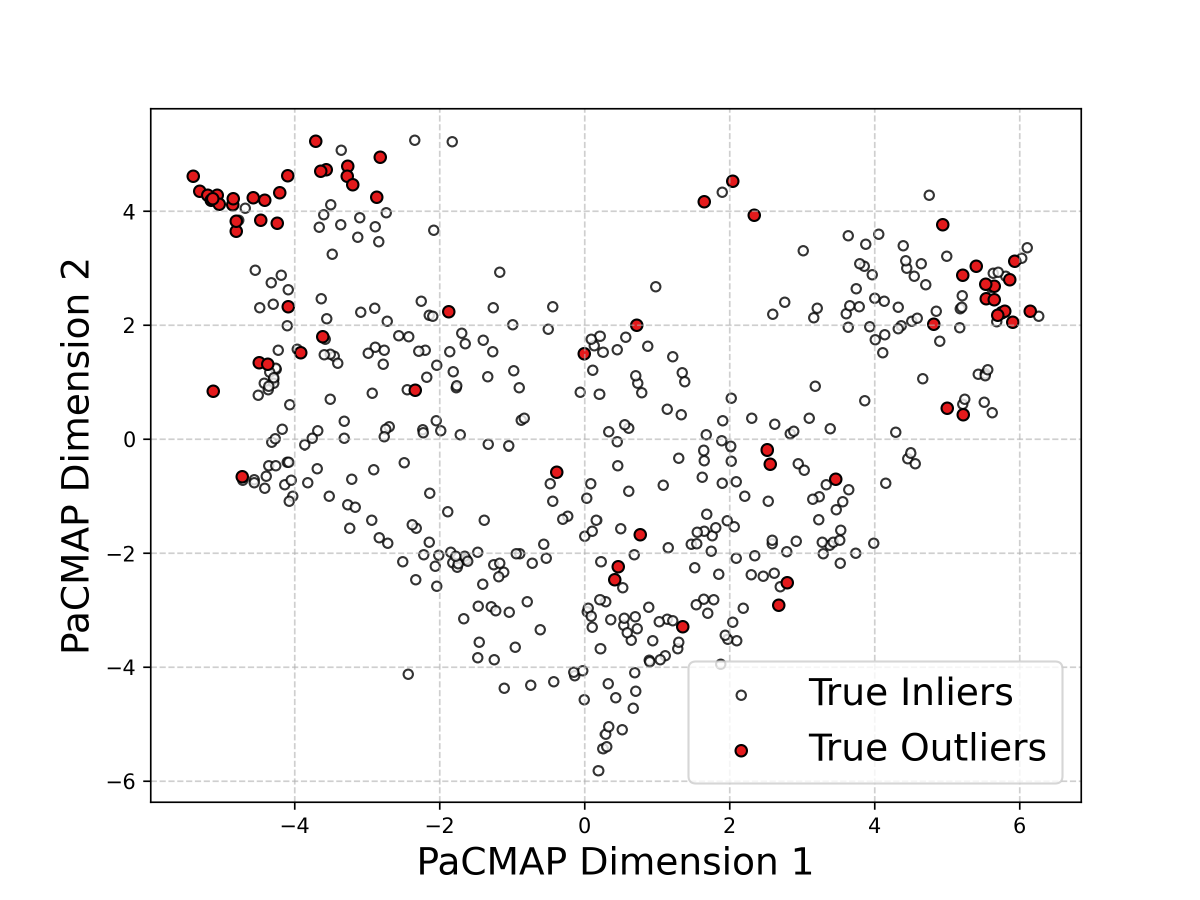}
  \caption{PaCMAP visualization of the Arrhythmia dataset; It shows a
    complex, unstructured distribution.}
  \label{fig:pacmap_arrhythmia}
\end{figure}

The Arrhythmia dataset, where our method also achieved the top score
(AUC: 0.813), presents a starkly different picture
(Fig.~\ref{fig:pacmap_arrhythmia}). The data exhibits no discernible
cluster structure; inliers and outliers are intermingled, forming a
single, sparse, and complex cloud in high-dimensional space.

This ``unstructured'' nature poses a significant challenge for many
algorithms. The success of our Two-Stage LKPLO in this scenario is
particularly noteworthy, as it demonstrates that our method is not
merely a cluster-based detector. Instead, the global kernel PCA stage
appears to be capturing high-order, non-linear correlations within the
data that are not visually apparent in a 2D embedding. The subsequent
``clustering'' stage may then be partitioning this complex feature
space into regions of differing statistical properties, allowing the
local projection-based scoring to identify subtle deviations that
constitute anomalies. The victory here highlights our method's
versatility: it excels on clearly structured data like Optdigits by
modeling the clusters, and on unstructured data like Arrhythmia by
leveraging the power of its kernelized, localized feature analysis.

\subsection{Ablation Study: The Synergy of Kernelization and Localization}

To empirically validate our architectural design choices, we conducted
an ablation study to disentangle the individual contributions of the key
components within our framework. We compare our full model, Two-Stage
LKPLO, against its simpler variants by progressively removing its core
components:

\begin{itemize}
\item \textbf{KPLO (Kernel, Global)}: This variant omits the
  localization stage (Stage 2 clustering), applying the kernelized
  projection search globally. Its purpose is to test the necessity of
  the localization strategy.
\item \textbf{PLO (Linear, Global)}: This variant omits both the
  kernelization (Stage 1) and the localization, operating directly in
  the original data space. It serves as a baseline to evaluate the
  impact of the non-linear mapping.
\end{itemize}

We perform this study on two synthetic datasets, Inside--Outside and
Vowels, each chosen to isolate and highlight the importance of a
specific architectural component. The performance, evaluated via
5-fold cross-validation and optimized with Optuna for each model, is
summarized in Table~\ref{tab:ablation_study}.

\begin{table}[t]
  \centering
  \caption{Ablation study results (mean ROC AUC $\pm$ std.\ dev.) on
    two synthetic datasets. The performance consistently improves as
    components are added, demonstrating the synergy of kernelization
    and localization.}
  \label{tab:ablation_study}
  \begin{tabular}{lcc}
    \toprule
    \textbf{Model Variant} & \textbf{Inside--Outside} & \textbf{Vowels} \\
    \midrule
    \textit{Baseline:} & & \\
    PLO (Linear, Global) & 0.446 $\pm$ 0.065 & 0.756 $\pm$ 0.074 \\
    \midrule
    \textit{+ Kernelization:} & & \\
    KPLO (Kernel, Global) & 0.513 $\pm$ 0.054 & 0.817 $\pm$ 0.057 \\
    \midrule
    \textit{+ Localization (Full Model):} & & \\
    \textbf{Two-Stage LKPLO} & \textbf{0.788 $\pm$ 0.141} & \textbf{0.919 $\pm$ 0.052} \\
    \bottomrule
  \end{tabular}
\end{table}

The Inside--Outside dataset, characterized by its highly non-linear
and non-convex structure, clearly demonstrates the importance of the
kernel mapping. As shown in Table~\ref{tab:ablation_study}, PLO, which
operates linearly, fails completely on this task with a near-random
AUC of 0.446. The introduction of kernelization in KPLO yields only a
marginal improvement to 0.513, but the model still fails to capture
the underlying structure. This indicates that while kernelization is a
necessary step to handle the non-linearity, it is not sufficient on
its own.

The crucial role of our two-stage localization strategy is highlighted
by the results on both datasets. On the Inside--Outside data, the
performance dramatically increases from 0.513 (KPLO) to 0.788
(Two-Stage LKPLO). This leap can be attributed to the clustering
stage, which successfully separates the inlier ring from the internal
outlier cluster, allowing each to be evaluated in its own local
context; this is a task that a global model fundamentally cannot
perform.

This finding is further corroborated by the Vowels dataset, a
canonical example of multi-modal data. Here, the ablation results show
a clear, stepwise improvement: PLO (0.756) $\rightarrow$ KPLO (0.817)
$\rightarrow$ Two-Stage LKPLO (0.919). The substantial performance
gain achieved by introducing the localization stage confirms that this
component is essential for effectively handling data with multiple
distinct inlier clusters.

Overall, this ablation study provides compelling evidence that the
superior performance of our proposed method arises not from a single
component but from the synergistic combination of kernelization and
localization. Kernelization is vital for adapting to non-linear
manifolds, while localization is indispensable for resolving
multi-modal structures. By integrating both, the Two-Stage LKPLO
architecture offers a powerful and flexible solution for a class of
complex outlier detection problems that are intractable for its
constituent parts alone.

\section{Conclusion}

In this paper, we introduced a novel outlier detection method,
Two-Stage Localized Kernel PLO (Two-Stage LKPLO), which
synergistically combines a generalized, loss-based measure of
outlyingness (PLO) with a two-stage architecture involving kernel PCA
and subsequent clustering-based localization. This integrated approach
is designed to effectively handle complex data exhibiting both
non-linear and multi-modal structures.

Comprehensive experiments, conducted using 5-fold cross-validation
with automated hyperparameter optimization, rigorously evaluated this
design. The proposed method achieved state-of-the-art or highly
competitive performance on several challenging datasets, most notably
Optdigits, Arrhythmia, and Vowels. The strong performance on Optdigits
validates our core hypothesis that the two-stage architecture
represents a powerful strategy for multi-cluster data.

Furthermore, our ablation study provided compelling evidence for the
synergistic effectiveness of our design choices. It empirically
demonstrated that both the kernelization stage and the localization
stage are indispensable components, each contributing significantly to
the overall performance on data with non-linear and multi-modal
characteristics, respectively. The consistent superiority of the
SVM-like loss over the classical Robust-Z metric further highlights
the fundamental advantage of our flexible PLO framework, which lies in
moving beyond fixed statistical metrics.

While no single method proves universally superior, our work
contributes a powerful new algorithm for a specific and challenging
class of outlier detection problems. More broadly, it offers a
flexible framework and concrete architecture that underscore the
importance of hybrid, multi-stage approaches in modern outlier
detection. Future work will focus on exploring more advanced
clustering algorithms to further enhance performance on complex
multi-modal data and on developing methods for the automated selection
of optimal loss functions. Additionally, while our current
implementation utilizing the RBF kernel is designed for continuous
numerical features, extending this framework to effectively handle
discrete or categorical data represents an important direction for
broadening its real-world applicability.

\bibliographystyle{ieicetr}

\appendix
\section{Evaluation with Logistic Loss Variant}
To further investigate the flexibility of the PLO framework, we
implemented and evaluated a third variant of our Two-Stage LKPLO using
a logistic loss function.

\subsection{Formulation of Logistic Loss}
The logistic loss is defined as:
\begin{equation}
\mathcal{L}_{\text{Logistic}}(\bm{u}, \bm{f}, \bm{F}'_k) = \log(1 + \exp(\bm{u}^T \bm{f}')),
  \label{eq:logistic_loss}  
\end{equation}
where the feature vectors are centered at the local cluster median,
i.e., $\bm{f}' = \bm{f} - \text{median}(\bm{u}^T \bm{F}'_{k})$.  This
loss can be interpreted as the negative log-probability of the point
belonging to the outlier class ($y = -1$) under a logistic model.

\subsection{Experimental Results}
We conducted the same 5-fold cross-validation experiment with
automated hyperparameter optimization for the logistic loss
variant. The results are summarized in Table~A$\cdot$1. Compared to
the SVM-like variant (Avg: 0.843), the logistic loss variant achieved
a lower overall performance (Avg: 0.810). This suggests that for the
datasets under consideration, the margin-based approach of the
SVM-like loss is a more effective choice.

\begin{table}[h]
  \label{tab:results_logistic}
  \centering
  \caption{ROC AUC scores (mean $\pm$ std.\ dev.) for the Two-Stage
    LKPLO variant using the logistic loss function. Results are based
    on 5-fold cross-validation with Optuna-based hyperparameter
    optimization.}
  \begin{tabular}{lc}
    \toprule
    \textbf{Dataset} & \textbf{Two-Stage LKPLO (Logistic)} \\
    \midrule
    Arrhythmia & 0.790 $\pm$ 0.032 \\
    Cardio     & 0.860 $\pm$ 0.034 \\
    Glass      & 0.898 $\pm$ 0.031 \\
    Ionosphere & 0.812 $\pm$ 0.021 \\
    Letter     & 0.752 $\pm$ 0.040 \\
    Optdigits  & 0.839 $\pm$ 0.112 \\
    Pima       & 0.688 $\pm$ 0.042 \\
    Vertebral  & 0.633 $\pm$ 0.136 \\
    Vowels     & 0.902 $\pm$ 0.038 \\
    WBC        & 0.931 $\pm$ 0.052 \\
    \midrule
    \textbf{Average} & 0.810 $\pm$ 0.096 \\
    \bottomrule
  \end{tabular}
\end{table}
  
\profile{Akira Tamamori}{He received his B.E., M.E., and D.E. degrees
  from Nagoya Institute of Technology, Nagoya, Japan, in 2008, 2010,
  2014, respectively.  From 2014 to 2016, he was a Research Assistant
  Professor at Institute of Statistical Mathematics, Tokyo,
  Japan. From 2016 to 2018, he was a Designated Assistant Professor at
  the Institute of Innovation for Future Society, Nagoya University,
  Japan. From 2018 to 2020, he was an lecturer at Aichi Institute of
  Technology, Japan.  He has been an Associate Professor at Aichi
  Institute of Technology since 2020. He is a member of the Institute
  of Electronics, Information and Communication Engineers (IEICE), the
  Information Processing Society of Japan (IPSJ), the Acoustical
  Society of Japan (ASJ), and Asia-Pacific Signal and Information
  Processing Association (APSIPA).}

\end{document}